\documentclass[letterpaper]{article} 
\usepackage{aaai24}  
\usepackage{times}  
\usepackage{helvet}  
\usepackage{courier}  
\usepackage[hyphens]{url}  
\usepackage{graphicx} 
\usepackage{natbib}  
\usepackage{caption} 
\usepackage{algorithm}
\usepackage{algorithmic}

\usepackage{amsmath}
\usepackage{amsfonts}
\usepackage{booktabs}
\usepackage{tabularx}
\usepackage{float}

\usepackage{lipsum}
\usepackage{amssymb}
\usepackage{siunitx}
\usepackage{tcolorbox}
\usepackage{tikz}

\usepackage{newfloat}
\usepackage{listings}
\DeclareCaptionStyle{ruled}{labelfont=normalfont,labelsep=colon,strut=off} 
\floatstyle{ruled}
\newfloat{listing}{tb}{lst}{}
\floatname{listing}{Listing}
\nocopyright

\title{Multimodal Contrastive In-Context Learning}
\author{
    Yosuke Miyanishi\textsuperscript{\rm 1,2}, Minh Le Nguyen\textsuperscript{\rm 1}\\
}
\affiliations{
    \textsuperscript{\rm 1}Japan Advanced Institute of Science and Technology\\
    \textsuperscript{\rm 2}CyberAgent Inc.\\
yosuke.miyanishi@jaist.ac.jp, nguyenml@jaist.ac.jp

}

\begin{document}

\maketitle

\begin{abstract}
The rapid growth of Large Language Models (LLMs) usage has highlighted the importance of gradient-free in-context learning (ICL). However, interpreting their inner workings remains challenging. This paper introduces a novel multimodal contrastive in-context learning framework to enhance our understanding of ICL in LLMs. First, we present a contrastive learning-based interpretation of ICL in real-world settings, marking the distance of the key-value representation as the differentiator in ICL. Second, we develop an analytical framework to address biases in multimodal input formatting for real-world datasets. We demonstrate the effectiveness of ICL examples where baseline performance is poor, even when they are represented in unseen formats. Lastly, we propose an on-the-fly approach for ICL (Anchored-by-Text ICL) that demonstrates effectiveness in detecting hateful memes, a task where typical ICL struggles due to resource limitations. Extensive experiments on multimodal datasets reveal that our approach significantly improves ICL performance across various scenarios, such as challenging tasks and resource-constrained environments. Moreover, it provides valuable insights into the mechanisms of in-context learning in LLMs. Our findings have important implications for developing more interpretable, efficient, and robust multimodal AI systems, especially in challenging tasks and resource-constrained environments. 
\end{abstract}
\section{Introduction}
Upon the explosive usage of the Large Language Model (LLM), in-context learning (ICL) characterizes LLM's reasoning process. Understanding its optimization mechanism is critical for reliable, evidence-based decision-making. Previous works have shown that LLMs could optimize the attention weights in the gradient-free inference. The research scope, however, is mostly limited to simple problems like linear regression or word-level natural language inference.\\
The recent advances in multimodal LLM present us with more challenges. First, in addition to the linguistic format dependencies, which seem trivial to humans, multimodal ICL involves arbitrarily formatted multiple modalities. Although the research community proposes many approaches for solutions in different contexts, the impact of the multimodal ICL input formatting remains elusive. Second, exploring effective in-context examples is demanding due to the limited source of high-quality multimodal datasets compared to those of single modality.\\
To achieve a deeper understanding of LLM, as the gradient descent hinted at attention-based optimization, the existing gradient-based learning method could help interpret how it optimizes in ICL. Specifically, Contrastive Learning (CL), typically used for modality encoders and classic language models, could guide the model in mapping semantically similar inputs to a similar location in feature space. Based on the previous theoretical findings about the equivalence of LLM's learning process and CL, we show that CL helps interpret how LLM understands multimodal ICL semantics under unseen input formatting and/or resource shortage.
Our contribution could be summarized as follows:
\begin{enumerate}
    \item We propose a first CL-based interpretation of ICL in multimodal settings, suggesting that the semantically similar ICL examples trigger the representational shift dependent on the problem settings.
    \item We propose a CL-based analytical framework for the bias of multimodal input formatting and show that semantically similar ICL examples could be helpful in challenging tasks even when presented in an unseen format.
    \item We propose Anchored-by-Text ICL, an \textit{on-the-fly} inference in which LLM first generates the ICL example and then performs the inference using the generated example as an anchor for extracting the input-label relationship. This approach has shown effectiveness in resource-limited settings.
\end{enumerate}

\section{Related Work}
In these few years, LLMs have been widely adapted to natural language processing \cite{zhaoSurveyLargeLanguage2023}, showing remarkable in-context learning (ICL) performance \cite{brownLanguageModelsAre2020} with up to a few examples and without gradient-based training. Massive work has tested their multimodal capabilities \cite{zhangMMLLMsRecentAdvances2024} centered on vision and language as a step toward general-purpose agents.

\subsection{Interpreting Inner Workings}
To achieve Trustworthy AI \cite{thiebesTrustworthyArtificialIntelligence2021}, understanding how LLMs achieve high ICL performance is imminent. Various interpretations have been proposed to obtain theoretical and empirical grounding behind ICL. Typically, the interpretation studies hire a specific algorithm to interpret the dynamics of LLM's representations: for example, Bayesian inference \cite{xieExplanationIncontextLearning2022a}, kernel regression \cite{hanInContextLearningLarge2023}, latent variable model \cite{wangLargeLanguageModels2023}, algorithm selector \cite{liTransformersAlgorithmsGeneralization2023}, multi-state RNN \cite{orenTransformersAreMultiState2024}, and gradient descent \cite{vonoswaldTransformersLearnInContext2023,daiWhyCanGPT2023a}. Although these studies covered extensive theoretical aspects, most empirical findings are limited to simple problems like linear modeling or simple NLP tasks, let alone multimodal settings.\\
To demystify LLM's remarkable multimodal ICL capabilities, its training and evaluation procedure should be a clue. Most LLMs are trained to maximize the predicted probability of the tokens in the training datasets \cite{shlegerisLanguageModelsAre2024}. In multimodal problems, non-language information (e.g., image) is encoded as captioned text (e.g. \citet{miyanishiCausalIntersectionalityDual2024}) or soft prompt (e.g. \citet{bulatLASPTexttoTextOptimization2023}). At inference time, ICL frameworks mostly anchor the semantically similar examples to the test input \cite{liuWhatMakesGood2022,wangLearningRetrieveInContext2024,liUnderstandingInContextLearning2023}, making it intuitive to hypothesize that the distance between ICL example and test input plays a crucial role in ICL. Here, we formally and empirically show that multimodal input distance, coded during the LLM's training procedure, plays an crucial role in understanding the ICL inputs.
To provide such an distance-oriented view of ICL, Contrastive Learning (CL) \cite{le-khacContrastiveRepresentationLearning2020} could play an pivotal role. CL was initially developed as an unsupervised approach for training data distribution, and then \citet{khoslaSupervisedContrastiveLearning2020} introduced supervised CL for labeled datasets. Before the paradigm shift to generative models, CL was a major pre-training objective of main-stream language models based on a Transformer \cite{vaswaniAttentionAllYou2017} encoder like BERT \cite{devlinBERTPretrainingDeep2019}. In the LLM era, its main application is multimodal (e.g. vision and language) alignment \cite{huBLIVASimpleMultimodal2024}. CL is rather straightforward for capturing the cross-input semantics since it is designed to map the inputs to the feature space based on conceptual similarity.\\
Recently, \citet{renIncontextLearningTransformer2023} has theoretically analyzed the equivalence of ICL and supervised CL without negative examples and has shown its validity in simple mathematical problem-solving. In addition, we extend the analysis to the multimodal real-world datasets and propose that the semantically similar ICL examples trigger the representational shift in LLMs.

\subsection{Input Formatting}
Prompt engineering \cite{bozkurtGenerativeAIPrompt2023} has tackled the optimization of the instruction and task description. In addition to the textual information, multimodality \cite{wangReviewLargeVision2023} poses a new challenge - how LLMs could understand the interleaved inputs of multiple information sources. Focusing on the image-text relationship, the most straightforward format is an image followed by a single instruction (e.g., a single visual question-answering entry), targetted by the most state-of-the-art multimodal models like LLaVA \cite{liuVisualInstructionTuning2023}. Another popular format is multi-turn conversation \cite{fengMMDialogLargescaleMultiturn2023,morganMultimodalDatasetReal2023}, with which the model should recognize at least the two lines of text interleaved by two images. Recent studies have tackled this problem with tailored pre-training protocol \cite{zhengMiniGPT5InterleavedVisionandLanguage2023} and/or instruction tuning \cite{liOtterMultiModalModel2023,tangCoDi2InContextInterleaved2023}. In line with these works, this paper quantitatively shows how the unseen format biases the LLM's comprehension of the ICL example, and that semantics-formatting balance works differently for the different tasks.

\subsection{Resource Shortage}
Like the limited vision-and-language ability of humans with less visual experience \cite{lopez-barrosoImpactLiteracyFunctional2020,mamusLackVisualExperience2023}, the resource shortage is a significant challenge for vision-oriented models \cite{baiSequentialModelingEnables2023}. Since typical ICL involves example selection from training subset of a given task, task-specific multimodal resources, like hateful memes detection datasets \cite{kielaHatefulMemesChallenge2020,gomezExploringHateSpeech2020}, constrains the ICL performance. 
One approach to this problem is to let the LLMs generate ICL examples for their own usage. For example, \citet{wangLearningRetrieveInContext2024} framed this problem into retrieval, and  \citet{coda-fornoMetaincontextLearningLarge2023a} has shown that LLMs  can perform meta-learning via ICL. Notably, in some cases like hateful memes, forcing state-of-the-art LLMs to generate \textit{positive} examples is challenging for safety reasons. This paper shows that LLM-generated \textit{negative} examples shift the model's representation, and mitigate this positive example constraint.

\section{Preliminaries}
\subsection{Learning Objective of Generative Transformers}
Transformer's self-attention layer of depth $d$ maps input document $D$ to query $Q$, key $K$, value $V$ with corresponding weight matrix $W$. An layer is written as:
\begin{equation}\label{eq1}
    \begin{aligned}
        Q&=W_QD,K=W_KD,V=W_VD\\
        Self&Attn(Q,K,V)=SoftMax(\frac{QK}{\sqrt{d}})V
    \end{aligned}
\end{equation}
In case of generating the answer $a$ for a set of the documents $D_{icl}=\{D_{query},D_{ex}\}$ consisting of the query $D_{query}$ with the ICL example $D_{ex}$, the predicted most probable answer $\hat{y}$ is obtained as:
\begin{equation}\label{eq2}
    \hat{y}=\mathop{\mathrm{argmax}}_y\ p(y|SelfAttn(D_{icl}))
\end{equation}
\subsection{ICL and CL}
CL typically utilizes contrastive loss \cite{hadsellDimensionalityReductionLearning2006} with which  the document pair $(D_1, D_2)$ is mapped to the representation space with the guidance of a binary $y_c$ (1 suggests that the documents are in a specific category, 0 otherwise). Given a distance function $dist(\cdot,\cdot)$,  the loss $\mathcal{L}$ with hyperparameter $\epsilon$ is defined as:
\begin{equation} \label{eq:3}
    \begin{aligned}
        \mathcal{L}(D_1, D_2)&=y_cd_{D_{1/2}}+(1-y_c)max(\epsilon-d_{D_{1/2}},0)\\
        where\ d_{D_{1/2}}&=dist(D_1, D_2)
    \end{aligned}
\end{equation}

In inference time, the learned function $f_{CL}$ maps the new input $D_{test}$ to the representation space, and a dedicated function $f_{proj}$ projects that representation to $\hat{y}$.
\begin{equation}\label{eq4}
    \hat{y}=f_{proj}(f_{CL}(D_{test}))
\end{equation}

\citet{renIncontextLearningTransformer2023} has shown that ICL could be seen as CL without negative examples. They suggested that a self-attention layer could be seen as a contrastive learner. More specifically, with the help of a kernel function $\phi$, a single layer minimizes the distance between two different augmentations $\hat{x}_{K}, \hat{x}_{V}$ of an identical training data point's representation $h$.
\begin{equation}\label{eq5}
    \begin{aligned}
        \hat{x}_K&=W\phi(W_Kh)\\
        \hat{x}_V&=W_Vh\\
        \mathcal{L}(\hat{x}_K,\hat{x}_V)&=d_{\hat{x}_{K/V}}
    \end{aligned}
\end{equation}

Note that the category label $y_c$ is omitted for the absence of the negative class. After the input passes through a single model layer, it gets the new representation $h'$, embeds it to the same feature space using the query weight $W_Q$, and obtains the inference output $\hat{y}$ using the updated weight $\hat{W}$.
\begin{equation}\label{eq6}
    \begin{aligned}
        \hat{W}&=W-\eta\Delta\mathcal{L}\\
        where&\ \Delta\mathcal{L}=\frac{\partial \mathcal{L}}{\partial W}\\
        \hat{y}&=\hat{W}x^{test}\\
        where&\ x^{test}=\phi(W_Qh')
    \end{aligned}
\end{equation}
Hereafter, we omit the learning rate $\eta$ for brevity. Since the weight update $\Delta \mathcal{L}$ is a function of key-value distance, we denote the update as $\Delta \mathcal{L}(K,V)$, and its resulting (ICL-optimized) weight as $W_{icl}$. This paper factorizes the real-world learning process and empirically shows its significance.

\subsection{Mixed Effect Model}
Mixed effect model \cite{singmannIntroductionMixedModels2019} has been proposed to disentangle the dual effects of the variables within the same model. Specifically, in observation $i$, the effect of some variables $X$ over the target variable $y_i$ is expected to be identical across all the observations (\textit{fixed effect}), and another variables $Z$ affect individual (group of) observation differently (\textit{random effect}). Linear mixed effect model could be formalized as:
\begin{equation}
    y_i=W_{X}X+W_{Z_i} Z_i
\end{equation}

For example, if we are to analyze the effect of a new teaching method on student performance across different schools in a city, the method should have a fixed effect since, in general, such a method aims for equal educational opportunities. In contrast, the school variable should have a random effect since each school must have a different educational policy. Note that various non-linear expressions of the mixed effect are proposed (e.g. \citet{hajjemMixedeffectsRandomForest2014,sigristLatentGaussianModel2023}), but we limit the scope to the linear model for brevity.

\subsection{ICL Example Selection}
In ICL, the example $D_{ex}$ is typically extracted from the training dataset or its subset $\bigcup D_{train}$ to obtain the closest example to the test input $D_{query}$.
\begin{equation}\label{eq7}
    D_{ex}=\mathop{\mathrm{argmin}}_{D_{train}}\ d_{D_{train}/D_{query}}
\end{equation}
We show the effectiveness of generating the example instead of selecting it and discuss how it is related to CL.

\section{Methodology}
\subsection{Outline of Our Method}
Fig.1 summarizes our method.
\begin{figure}[!ht]
    \begin{center}
    \includegraphics[scale=0.5]{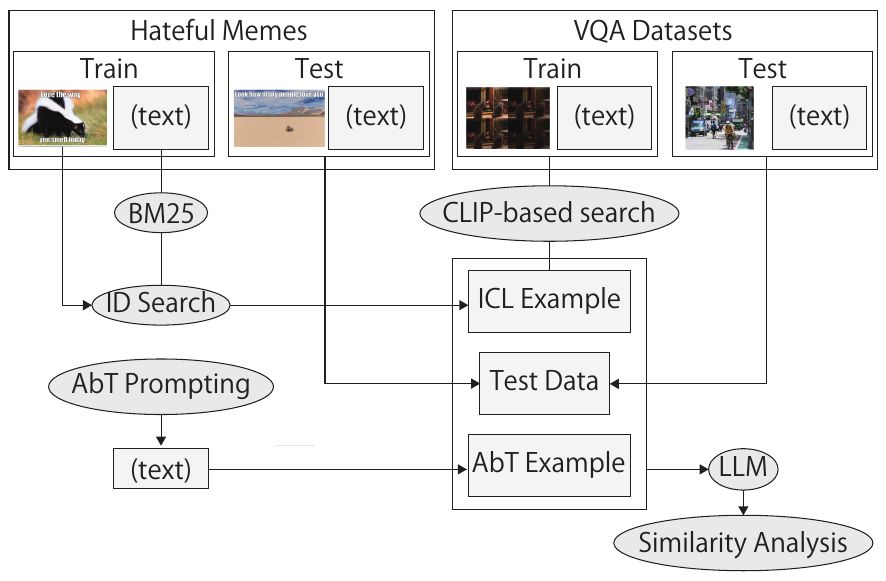} 

    \caption{Summary of the proposed method. Boxes represent data, while circles symbolize procedures and models. Images are taken from Hateful Memes \cite{kielaHatefulMemesChallenge2020} and MMBench \cite{liuMMBenchYourMultimodal2023} datasets.}
    \label{fig.1}
    \end{center}
\end{figure}
\subsection{Representational Shift Hypothesis}
In CL interpretation (Eq.5-6), key-value distance contributes to attention-based optimization in ICL. Since this interpretation only presupposes the interaction of key-value pair, we could extend it to the arbitrary set of test-time input fractions (e.g. instruction prompt $K_{inst}$ and the task given in zero-shot setting $V_{zsl}$). Since the zero-shot task consists of the instruction, the task, and LLM's prediction $pred$, and the model is trained to infer the latter tokens from the former, we propose that the distance among the zero-shot components affects the generation as follows:
\begin{equation}
    \begin{aligned}
        W_{zsl}&=W-\Delta W(K_{inst},V_{zsl})\\
        W_{pred}&=W_{zsl}-\Delta W(K_{zsl},V_{pred})
    \end{aligned}
\end{equation}

Similarily, the updated ICL weight $W_{icl}$ could be formalized as:
\begin{equation}
    \begin{aligned}
        W_{icl}&=W-\{\Delta W(K_{inst},V_{icl})+\Delta W(K_{icl},V_{zsl})\}\\
        W'_{pred}&=W_{icl}-\Delta W(K'_{zsl},V_{pred})
    \end{aligned}
\end{equation}
Assuming that the overall instruction affects each task equally $\Delta W(K_{inst},V_{zsl}) \simeq \Delta W(K_{inst},V_{icl})$, the weight (and resulting representation) \textit{shifts} towards example-task distance.
\begin{equation}
    \begin{aligned}
        W'_{pred}-W_{pred}&=\Delta W(K'_{zsl},V_{pred})-\Delta W(K_{zsl},V_{pred})\\
        &=\Delta W(K_{icl},V_{zsl})
    \end{aligned}
\end{equation}
In summary, the ICL example first affects the representation of the zero-shot task, and the prediction is affected via the task-prediction representational shift. To test whether this hypothesis is correct, we analyze the distance-distance relationship.

\subsection{Multimodal Input Formatting}
\subsubsection{Disentangling Format and Semantics}
The semantics of the bimodal inputs and their format are entangled yet different concepts.
Since CL aims to learn the inputs' similarity and variance, we could assume semantic similarity as its objective while formatting as a biasing factor. In other words, the formatting term $\mathcal{L}_{fmt}$ affects the actual loss $\mathcal{L}$ in parallel with its semantic term $\mathcal{L}_{sem}$.
\begin{equation}\label{eq8}
    \begin{aligned}
            \mathcal{L}&=\mathcal{L}_{sem}+\mathcal{L}_{fmt}\\
        \hat{W}&=W-(\Delta\mathcal{L}_{sem}+\Delta\mathcal{L}_{fmt})
    \end{aligned}
\end{equation}
Intuitively, within a single dataset, the second term consistently biases all the ICL examples (\textit{fixed effect}). In contrast, the first term should also reflect the variance of the individual test data points (\textit{random effect}). Therefore, when the model faces a test input with an unseen format, the model output for input $i$ should be interpreted as a mixed model.
\begin{equation}\label{eq9}
    \hat{y}_i=\{W-(\Delta\mathcal{L}_{sem}^i+\Delta\mathcal{L}_{sem}+\Delta\mathcal{L}_{fmt})\}x_i
\end{equation}
When the ICL format is the same with the training process, $\Delta\mathcal{L}_{fmt}=0$.

\subsubsection{Model Performance Analysis}
In the macroscopic view, the effect of the unseen format should be expressed as the impact of bias $b\in\{0,1\}$, and that of ICL example presence $e\in\{0,1\}$. Intuitively, the ICL examples have random effects due to the dependence on the content of each example. In contrast, the format bias should have a fixed effect, affecting the overall performance. Note that we use accuracy as the metric unless stated otherwise since all Visual Question Answering (VQA) datasets used in our analysis hire this metric. Together, the model accuracy $acc$ of a data subset $i$ could be modeled as:
\begin{equation}
    acc_i(b,e)=Wb+W_ie
\end{equation}
To analyze the impact across the models, the results of all the models are concatenated and the variables $b$ and $e$ are analyzed as an interaction term.

\subsubsection{Representation Analysis}
Since some of the widely used benchmarks like MMBench (\citet{liuMMBenchYourMultimodal2023}) require online submission for evaluation, which makes reproducible local evaluation challenging, we need the unsupervised approach. Under the our hypothesis, ICL is driven by the distance $d_{\cdot/\cdot}$ between the representation of the key $h_k$ and that of value $h_v$. In zero-shot VQA, the distance between question $h_q$ and answer $h_a$ would be the only clue to the model.  In contrast, the ICL example is concatenated to the question $h_{icl}=\{h_{ex},h_q\}$, leading to the shift in the feature space and, therefore, distance with the new answer $h_a'$.

In the spirit of the linear representation hypothesis (\citet{parkLinearRepresentationHypothesis2023}), we implement a linear mixed effect model. Specifically, the random effect is modeled as the linear weight $W_{random}$, and the fixed effect is introduced via the product of $h_{zcl}$ and the embedded index representing the model and the dataset with the weights  $W_{fixed}$. 
\begin{equation}\label{eq13}
    \begin{aligned}
        h_{icl}=(W_{random}+W_{fixed}I)h_{zsl}
    \end{aligned}
\end{equation}
Finally, we model the linear relationship between the query-answer distance matrix and the shifted query-new answer matrix.
\begin{equation}
    d_{h_{icl}/h_a'}=W d_{h_q/h_a}+W_0
\end{equation}
As a baseline, we use the model only with first term $h_{icl}=W_{random}h_{zcl}$, or a simple linear projection.

\subsection{Anchored-by-Text ICL}
\subsubsection{Generation Strategy}
Two major blockers must be addressed for on-the-fly ICL example generation on hateful memes. First, it requires text-image bimodal generation. Since only limited models (e.g. \citet{wuNExTGPTAnytoAnyMultimodal2023}) have such capability, we use the generated text as an \textit{anchor} to cause a representational shift, and therefore the prediction. Hereafter we call it anchored-by-text ICL (AbT ICL).\\
Second, most LLMs have safety limitations based on instruction tuning (\citet{bianchiSafetyTunedLLaMAsLessons2023}). Since bypassing such limitations is neither desirable nor sustainable, we let the model generate \textit{negative} examples. Together, given that document $D$ consists of text $T$ and image $I$ ($D=(T, I)$) with a binary label $y$ (0 for benign, 1 for hateful), our strategy is formalized as:
\begin{equation}\label{eq10}
    \begin{aligned}
        T_{icl}&=\mathop{\mathrm{argmax}}_{T}\ p(y=0|T,I_{query})\\
        D_{icl}&=\{T_{icl},I_{query}\}
    \end{aligned}
\end{equation}
In short, the model generates text that fits with a given image to compose a benign meme and uses that meme as a benign example. Baselines include zero-shot and one-shot detection. Fig.2 shows a representative prompt aiming for this goal.
\begin{figure}[t]
    \centering
    \footnotesize
    \begin{tcolorbox}[boxrule=0.2mm, width=0.99\columnwidth]
        \begin{tabular}{@{}p{0.97\columnwidth}@{}}
        \textbf{System:}\\
        You are a helpful language and vision assistant.
        
        \textbf{User:}\\
        \textcolor{blue}{\textless image in dataset\textgreater}\\
        Give me one caption that fits with this image.

        \textbf{Assistant:}\\        
        \textcolor{blue}{\{generated caption\}}\\
        
        \textbf{User:}\\
        In comparison with that caption, is the following caption hateful or benign? Answer with a single word.\\
        \textcolor{blue}{\{caption in dataset\}}\\
        
        \textbf{Assistant:}\\
        \textcolor{blue}{\{answer\}}
        \end{tabular}
    \end{tcolorbox}
    \caption{The representative Anchor-by-Text ICL prompt. The system prompt is truncated for illustrative purpose\footnote{Full version continues as: \textit{You are able to understand the visual content that the user provides, and assist the user with a variety of tasks using natural language.}}.}
    \label{fig:translation_prompt}
\end{figure}

\citet{quEvolutionHatefulMemes2023} introduced another workaround of using more general labels, which will be a part of our future work.

\subsubsection{Representation / Prediction Analysis}
Since hateful memes detection could be framed into binary classification in this experiment, we model the effect of the key-value distance (Eq. 13) over the predicted label $y$ on three learning types $lt$ (zero-shot $zsl$, ordinary ICL $icl$, and AbT ICL $abt$), and analyzed the difference of the weights $W$ and the intercept $W_0$ as an effect of the representational shift. For example, the effect of AbT over that of ordinary ICL could be formalized as:
\begin{equation}
    \begin{aligned}
        lt &\in \{zsl,icl,abt\}\\
        y_{lt}&=W^{lt} d_{h_{lt}/h_a}+W_0^{lt}\\
        y_{abt}-y_{icl}&=(W^{abt}-W^{icl})d_{h_{lt}/h_a}+(W_0^{abt}-W_0^{icl})\\
    \end{aligned}
\end{equation}
The weights $W^{lt}$ and $W_0^{lt}$ are estimated per layer dimension to perform the memory-efficient analysis.

\section{Experimental Settings}
\subsection{Shared Settings}
Experiments are conducted on a single NVIDIA A100 80GB GPU with Linux OS. Unless stated otherwise, all codes are in Python 3.9. Statistical arguments are based on a t-test and bootstrapping with 1,000 resamples. We run the models once with a random seed of 1987.

\subsection{Experiment I: Multimodal Input Formatting}
\subsubsection{Model}
To disentangle the effect of input semantics and that of the formatting, the subject model in this paper should 1) have the expected maximum capability of understanding the semantics and 2) is NOT trained or fine-tuned on a multi-image setting. We primarily focus on LLaVA \cite{liuVisualInstructionTuning2023} to satisfy this criterion. More specifically, we use two variants: \textit{LLaVA-Llama2} for its high performance of the linguistic backbone (\citet{touvronLlamaOpenFoundation2023}) and \textit{LLaVA 1.5} for its highest performance on vision-and-language tasks (\citet{liuImprovedBaselinesVisual2023}). 13 billion parameter models are used for memory constraints. We also use InternVL (1-8 billion) for their limited \footnote{https://github.com/OpenGVLab/InternVL/issues/419} yet tested multi-image capabilities by multi-image datasets like MMMU \cite{yueMMMUMassiveMultidiscipline2024}.\\
To select ICL examples most similar to test inputs, CLIP (\citet{radfordLearningTransferableVisual2021}, specifically HuggingFace \textit{clip-vit-large-patch14}) is used because of its relatively small computational cost and its high capability on similarity-related tasks (e.g., image aesthetics evaluation\footnote{https://laion.ai/blog/laion-aesthetics/}). We take the last layer as a representation for its high correspondence with the generated tokens despite the presence of highly competitive short-cutting \cite{dinJumpConclusionsShortCutting2024,fanNotAllLayers2024}.

\subsubsection{Dataset}
To cover various aspects of multimodal LLM's capabilities, we tested our approach with six VQA datasets, namely VQA v 2.0 (\citet{goyalMakingVQAMatter2017}), GQA (\citet{hudsonGQANewDataset2019}), VizWiz (\citet{gurariVizWizGrandChallenge2018}), TextVQA (\citet{singhVQAModelsThat2019}), MMBench (\citet{liuMMBenchYourMultimodal2023}), and MM-Vet (\citet{yuMMVetEvaluatingLarge2023}).

\subsubsection{Model Accuracy Analysis}
Practically, the presence of the  random and fixed effect ($z$ and $e$ in Eq. 13, respectively) is represented as a coefficient of the corresponding one-hot encodings. The performance of the mixed effect model is evaluated using the marginal/conditional R2 method (\citet{nakagawaGeneralSimpleMethod2013}). To maintain the experiment's integrity while utilizing a wide range of statistical tools, the R language's \textit{lmer} package is called from the Python environment via \textit{rpy2}\footnote{https://rpy2.github.io/doc.html} module.

\subsubsection{Representation Analysis}
The linear mixed model and the baseline linear model are implemented with PyTorch backend\footnote{https://pytorch.org/} and trained to maximize the cosine similarity between the representation via Pytorch Metric Learning package\footnote{https://kevinmusgrave.github.io/pytorch-metric-learning/} and AdamW optimizer (\cite{loshchilovDECOUPLEDWEIGHTDECAY2019}). We extract 1,000 samples from each dataset and hold out 20\% as a test set.

\subsection{Experiment II: AbT ICL}
Intuitively, the impact of AbT ICL may vary across datasets. The most influential scenario is 1) when the dataset size is small and suffers from high variance, making the example selection infeasible 2) when explicit and strong cross-modal interaction affects the dataset.\\
\citet{kielaHatefulMemesChallenge2020} curated the Hateful Memes Challenge dataset, which perfectly fits this experiment's criteria. Initially, \citet{laurenconOBELICSOpenWebScale2023} and \citet{zhaoMMICLEmpoweringVisionlanguage2023} have shown that ICL is not particularly effective unless the task is heavily tuned to the task. Moreover, \citet{heeExplainingMultimodalHateful2022} and \citet{miyanishiCausalIntersectionalityDual2024} theoretically and empirically showed that the cross-modal interaction embedded in the hateful memes detection problem is fully reflected in this dataset. We leave more experiments on hateful meme detection (\citet{gomezExploringHateSpeech2020}) and other tasks to future work.

\subsubsection{Model}
To comply with Experiment I, we use LLaVA-Llama2 in this experiment. For ICL example selection, we use BM25 algorithm \cite{robertsonOkapiTREC41996}.

\subsubsection{Dataset}
We focus on the Hateful Memes Challenge dataset \cite{kielaHatefulMemesChallenge2020} to test our framework in the context of complex multimodal interaction. Taking into account the presence of the \textit{image confounders} (two memes with identical text and different images, resulting in different labels), the one-shot experiment adopts the ICL examples with most similar texts (one meme from hateful, one meme from benign) in the labeled training set, and use the two confounders as a single set of ICL example. Since the data size is small, we use f1 score to see the precision-recall balance.

\section{Results \& Discussion}
\subsection{Experiment I: Multimodal Input Formatting}
\subsubsection{Motivation}
If the representational shift hypothesis is correct, the ICL examples could affect the prediction even if given in a format different from that of the training. The preliminary analysis shows that LLaVA \cite{liuVisualInstructionTuning2023}, a model not trained by multi-image datasets, can explain multiple images per prompt separately under some constraints (Supplementary Fig.1).

Based on this observation, our working hypothesis for Experiment I is that, although LLMs are heavily affected by the prompt format, they could interpret the semantics without solid inductive bias to some extent. We focus on ICL with a single example since we do not see any positive clue for further concatenation in the initial exploration.
\subsubsection{Performance}
Fig.3 and Supplementary Fig.4 summarize the performance of two LLaVA variants with or without the input of unseen format. Not surprisingly, LLaVA v1.5 outperforms v1 in all cases. Since the models are not trained with multiple-image datasets, the majority of the datasets show dropped performance in ICL. Interestingly, for LLaVA-Llama2, however, two image-text pairs boost the performance in some cases where the base performance is very low. This result supports the presence of semantics-based ICL, particularly when the task is challenging. 
\begin{figure}[!ht]
    \begin{center}
    \includegraphics[scale=0.5]{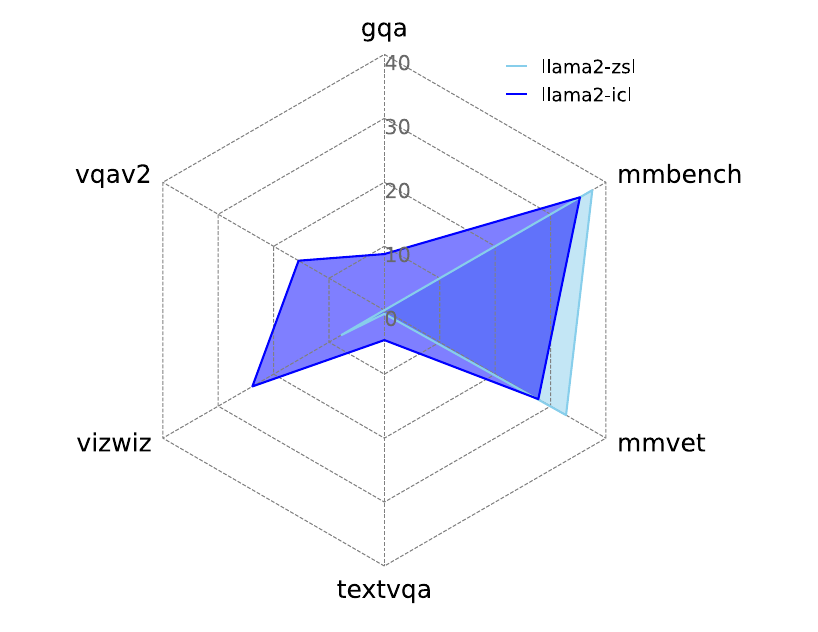} 

    \caption{Performance summary of LLaVA-Llama2. zsl and icl represent the corresponding learning type in the Methodology section.}
    \label{fig.3}
    \end{center}
\end{figure}
In the case of InternVL, ICL generally resulted in decreased performance, potentially because of its high performance and multi-image resource shortage (Supplementary Fig.2). To see whether the task difficulty affects this trend, we see the performance by the number of reasoning steps, typically seen as the difficulty metric, and is provided in the GQA dataset. Divided by this subcategory, ICL performs slightly better in the larger number of steps, in contrast to the dramatically dropped performance in the smaller number of steps (Table 1, Supplementary Fig.3). Together with LLaVA, these results suggest that the semantics dominate the challenging tasks, while the formatting is more critical in established ones.
\begin{table}[htbp]
    \centering
    \begin{tabular}{cccc}
        \toprule
        N Steps & N Samples & ZSL & ICL \\
        \midrule
        1-5 & 12,153 & $59.7 \pm 0.15$ & $52.5 \pm 0.31$ \\
        6-9 & 65 & $83.5 \pm 0.24$ & $84.6\pm 0.27$ \\
        \bottomrule
    \end{tabular}
    \caption{Impact of multi-image ICL in GQA for InternVL 1b. N steps indicates the number of inference steps. The numbers with error indicates accuracy(\%) in the corresponding setting.}
    \label{tab:multi-image-icl}
\end{table}

\subsubsection{Model Accuracy Analysis}
To quantify the impact of formatting and ICL examples, we model the linear mixed effect with or without the variables $\{z,e\}$, and the model variable $m$ (Table 2). In general, $m$ predominantly explains the accuracy variation, reflecting much higher performance for LLaVA 1.5. In $z$-$e$ comparison, $e$ has a slightly higher explanatory power, implying the significance of individual ICL example. This is further supported by the highest power of random effect when combined with $m$.
\begin{table}[htbp]
    \centering
    \small
    \setlength{\tabcolsep}{6pt}
    \begin{tabular}{@{}llcc@{}}
        \toprule
        \multicolumn{2}{c}{Variable} & \multicolumn{2}{c}{R\textsuperscript{2}*100} \\
        \cmidrule(lr){1-2} \cmidrule(lr){3-4}
        Fixed & Random & Fixed & Random \\
        \midrule
        m& m& $22.6 \pm 3.0$ & $52.0 \pm 8.8$ \\
        z& e& $0.3 \pm 0.1$ & $0.5 \pm 0.2$ \\
        m& e& $33.5 \pm 2.4$ & $33.6 \pm 2.5$ \\
        z& m& $0.2 \pm 0.1$ & $49.5 \pm 2.7$ \\
        comb  & comb  & $23.7 \pm 4.4$ & $53.7 \pm 8.8$ \\
        \bottomrule
    \end{tabular}
    \caption{Fixed and Random Effects.R\textsuperscript{2} values are multiplied by 100 for brevity. $m$ represents the model (LLaVA 1.5 or LLaVA-Llama2). $z$ and $e$ represents the formatting bias and the presence of ICL example, respectively. $comb$ represents the combined effect of the two variables in the same column. $R^2$ values are multiplied by 100 for brevity.}
    \label{tab:comparison}
\end{table}

\subsubsection{Representation Analysis}
First, to see if the representation of the ICL model's question-answer dinstance vector can be linearly mapped onto a zero-shot vector, we applied a simple linear probe to get moderate explanatory power with an $R^2$ value of $43.0 \pm 1.2$, suggesting the presence of such mapping. Next, we applied the high-dimensional mixed effect model (Eq. 16), resulting in a much higher R2 $59.2 \pm 2.3$. This result suggests that the representational shift in the presence of formatting bias could be mapped linearly. Next, we attributed the shifted representation to the original one together with the bias information (model and dataset, Table 3). The original score shows much higher than the bias binaries themselves, suggesting that those bias are interactive with model representation. In summary, these results suggest the presence of the linear mapping before/after the representational shift, and its effect could be seen as a mixed effect together with model and dataset. 
\begin{table}[h]
    \centering
    \begin{tabular}{lc}
        \toprule
        variable & coef*100 \\
        \midrule
        (Intercept) & $9.2 \pm 2.1$ \\
        \midrule
        mm-vet & $-0.75 \pm 0.7$ \\
        mmbench & $2.81 \pm 0.7$ \\
        textvqa & $2.1 \pm 0.6$ \\
        vizwiz & $0.16 \pm 0.7$ \\
        vqav2 & $-0.12 \pm 0.6$ \\
        \midrule
        model & $-0.39 \pm 0.4$ \\
        \midrule
        original score & $70.33 \pm 5.9$ \\
        \bottomrule
    \end{tabular}
    \caption{Mapping for the representational shift with bias information.}
\end{table}


\subsection{Experiment II: AbT ICL for Hateful Memes}
\subsubsection{Performance}
In comparison to the zero-shot setting, ICL significantly dropped the performance (Table 4). In contrast, AbT slightly improves the performance. These results suggest the capability of AbT in the absence of effective ICL examples. Further exploration for ineffective ICL problems will be the part of our future works. 
\begin{table}[h]
    \centering
    \begin{tabular}{lc}
        \toprule
        setting & f1*100 \\
        \midrule
        ZSL & $61.4 \pm 0.5$\\
        ICL & $58.5 \pm 0.9$\\
        AbT & $62.2 \pm 0.3$\\
        \bottomrule
    \end{tabular}
    \caption{Hateful memes detection performance.}
\end{table}
\subsubsection{Representation / Prediction Analysis}
We applied a linear probe between the distance vector and the predicted label to test the explanatory power of the key-value distance over the model prediction. This resulted in a moderate AUC of $75.6\pm0.90$, further supporting the contribution of key-value distance to the generation. Next, we extract each dimension's weight to see how $d$ shifts across the three settings (Fig.4). Interestingly,  AbT representation is close to that of ZSL, irrelevant of the labels, while ICL representation is distant. This result suggest that closer representation shift affects positively in case of hatetul memes detection.
\begin{figure}[!ht]
\begin{center}
\includegraphics[scale=0.7]{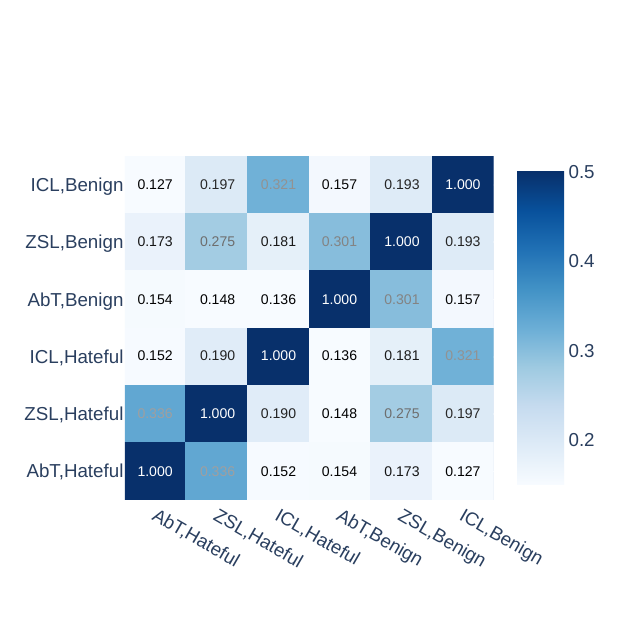} 

\caption{Representational shift across the learning type. Suffixes 0 and 1 represents the weights for benign and hateful}
\end{center}
\end{figure}

\subsection{Discussion}
Upon the previous pioneering study by \cite{renIncontextLearningTransformer2023}, our study on Multimodal Contrastive In-Context Learning (MCICL) has yielded several important findings that contribute to our understanding of in-context learning in LLMs.
\begin{enumerate}
    \item \textit{Representational Shift Hypothesis}: The representation analysis of two experiments supports our hypothesis. This finding provides insights into the mechanisms underlying ICL and suggests potential avenues for further optimization of ICL techniques.
    \item  \textit{Impact of Input Formatting}: Our results show that balancing the formatting and semantics of ICL inputs plays a crucial role in ICL performance.
    \item \textit{Anchored-by-Text ICL}: The proposed Anchored-by-Text ICL approach demonstrates effectiveness in resource-constrained hateful meme detection, important implication for real-world LLM applications.
\end{enumerate}

\subsection{Limitations and Future Work}
While our study provides valuable insights, there are several limitations and future research directions that warrant further investigation. Importantly, our experiments focused on a limited set of multimodal datasets and model architectures. Future work should explore the broader range of multimodal tasks and models, including but not limited to, multi-image tasks such as MMMU \cite{yueMMMUMassiveMultidiscipline2024} and missing modality problem \cite{wangMultiModalLearningMissing2023a,zhaoMissingModalityImagination2021}. In addition, whether the representational shift \textit{causes} the outcome variance is still elusive. One idea is to hire a mechanistic approach, such as path patching \cite{hannaHowDoesGPT22023,goldowsky-dillLocalizingModelBehavior2023}. Training phase mechanisms such as grokking or double descent \cite{daviesUnifyingGrokkingDouble2022} should also be part of the research scope.

\section{Conclusion}
MCICL enhances our understanding of in-context learning in LLMs by leveraging contrastive learning principles and addressing multimodal input challenges. It demonstrates improved performance in various scenarios, particularly in challenging settings.\\
Our work provides valuable insights but also highlights the need for continued research in multimodal learning complexity. MCICL opens new avenues for enhancing LLM capabilities in multimodal settings, contributing to more robust, efficient, and responsible AI systems.\\
As AI continues to evolve, approaches like MCICL will be crucial in creating more adaptable, interpretable, and effective multimodal AI systems for diverse real-world applications.


\section{Acknowledgments}
TBD

\bibliography{aaai24}

\appendix
\section{Appendix}
\subsection{Supplementary Figures}
\begin{figure}[ht!]
    \centering
    \footnotesize
    \begin{tcolorbox}[boxrule=0.2mm]
        \begin{tabular}{@{}p{0.97\columnwidth}@{}}
        \textbf{User:}\\
        Can you describe this image?\\
        \includegraphics[scale=0.5]{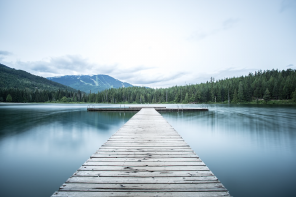}

        \textbf{Assistant:}\\
        (A)
        
        \textbf{User:}\\
        Can you describe this image?\\
        \includegraphics[scale=0.5]{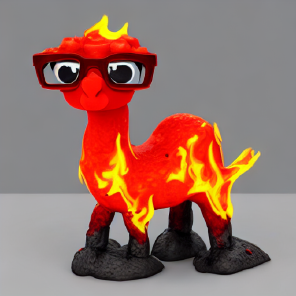}
        
        \textbf{Assistant:}\\
        (B)\\
        \hline\\
        LLaVA-LLama2:
        (A) A lake.
        (B) A red toy standing next to a lake.
        
        LLaVA 1.5:
        (A) A lake.
        (B) A red toy.
        \end{tabular}
    \end{tcolorbox}
    \caption{Comparison of model responses to two-image inputs. Images obtained from official LLaVA repository. The LLaVA response is truncated for brevity (see our repository for the full output). The main difference is in the description (B) of the second image. LLaVA-LLama2 explained that the red toy stands beside a lake, confusing the two images. LLaVA 1.5, on the other hand, gave a description that only mentioned the content in the second image, suggesting that it could disentangle the two images.}
    \label{fig:model_comparison}
\end{figure}
\begin{figure}[!ht]
    \begin{center}
    \includegraphics[scale=0.5]{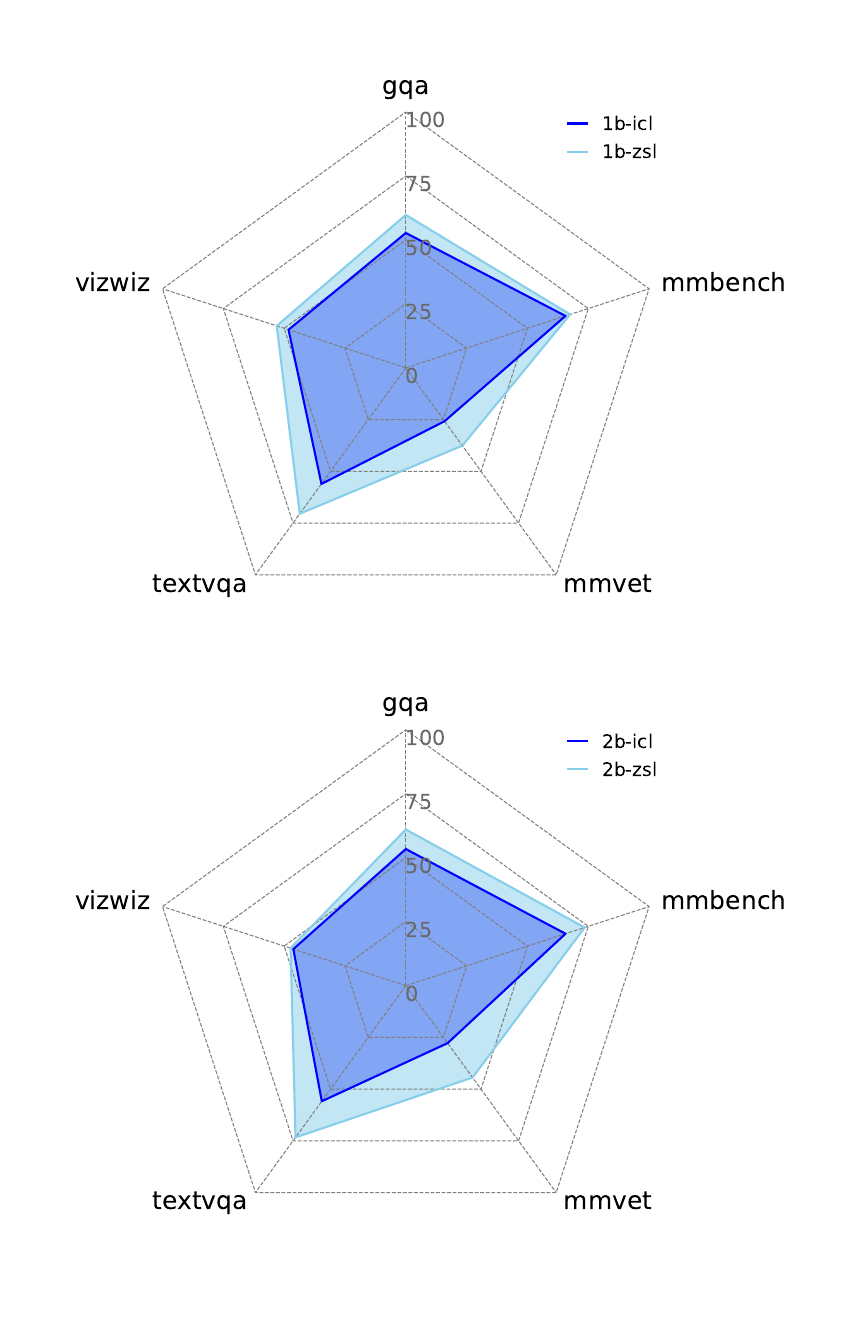} 
    \caption{Performance summary of InternVL. 1b and 2b indicates the number of model parameters.}
    \label{fig.s2}
    \end{center}
\end{figure}
\begin{figure}[!ht]
    \begin{center}
    \includegraphics[scale=0.5]{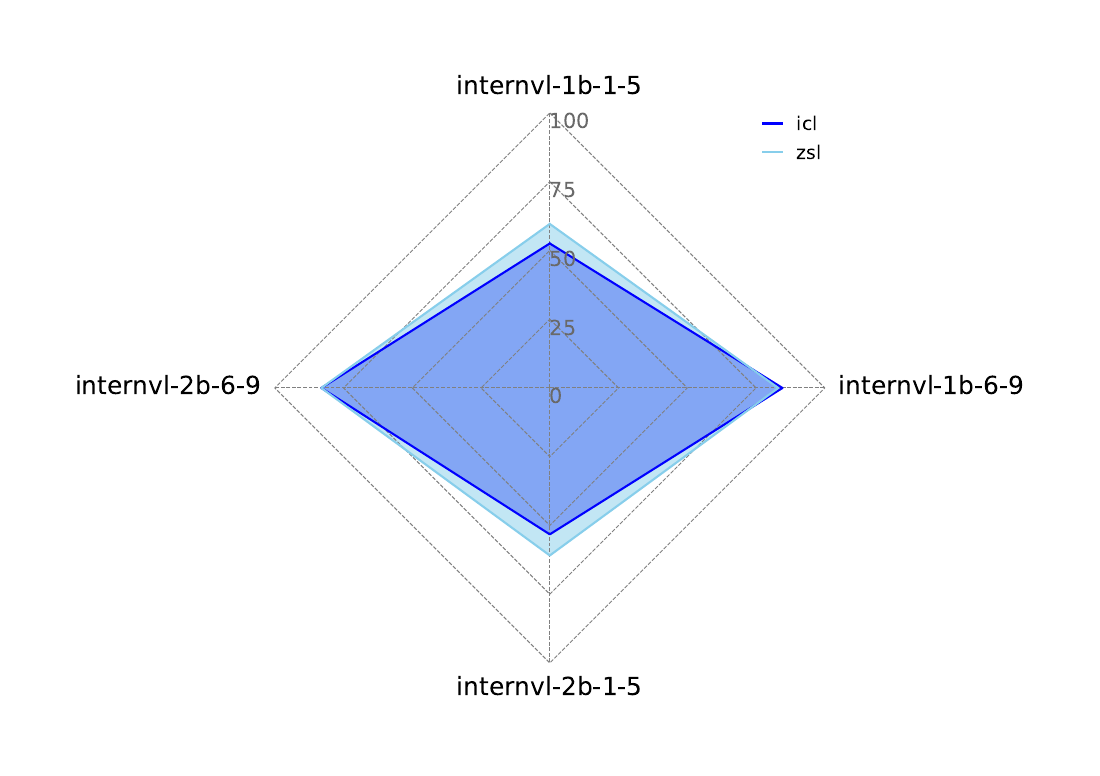} 
    \caption{GQA performance of InternVL by the number of inference steps.}
    \label{fig.s3}
    \end{center}
\end{figure}

        
        
        
        
\begin{figure}[!ht]
    \begin{center}
    \includegraphics[scale=0.5]{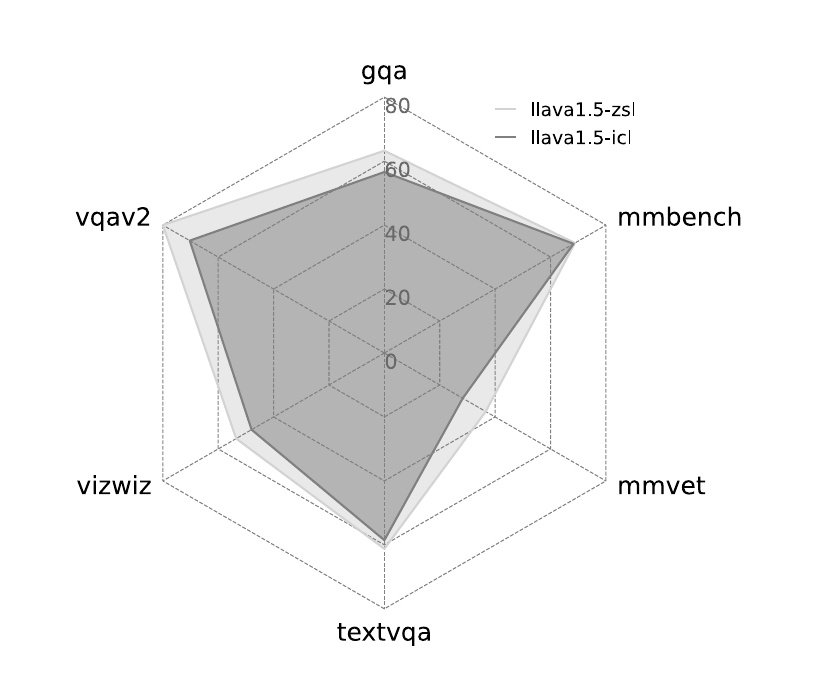} 

    \caption{Performance summary of LLaVA 1.5.}
    \label{fig.s4}
    \end{center}
\end{figure}

\subsection{Additional Discussion on Causality}
We leave the causal intervention to LLMs for future work. The nature of our framework, however, provides some causal explanation of the phenomena of interest, or the causality \textit{of the phenomena on the model}.
The causal effect could be helpful in quantitatively assessing how the phenomena of interest (e.g., unseen format, ICL example) affect the subject (LLM). For example, a widely used metric termed Average Treatment Effect (ATE) (\citet{rubinObjectiveCausalInference2008}) is defined as the average difference of outcome $y$ where the treatment $Z$ is given. Assuming binary treatment $Z\in\{Z_0,Z_1\}$, $ATE$ is formalized as:
\begin{equation}
    ATE=\mathbb{E}[y|Z_1]-\mathbb{E}[y|Z_0]
\end{equation}
Similarly to Eq.1, the causal effect on the prediction $y$ of the ICL example $e$ under the presence of unseen format bias $b$ in comparison with the zero-shot setting could be defined as the difference of the expected prediction between ICL $(b,e)=\mathbb{1}$ and zero-shot $(b,e)=\mathbb{0}$ settings.
\begin{equation}
    ATE_{macro}=\mathbb{E}[y|\mathbb{1},D_{icl}]-\mathbb{E}[y|\mathbb{0},D_{query}]
\end{equation}
Since the accuracy metric $acc$ is the ratio of correct prediction over the samples, $acc$ is identical to $\mathbb{E}[y]$, where $y$ is a binary for the correct prediction. Therefore, analyzing the accuracy difference provides us with insights into $ATE$.
\begin{equation}
    ATE_{macro}=acc(\mathbb{1})-acc(\mathbb{0})
\end{equation}
Similarly, the causal effect of ICL over the model on CL perspective is:
\begin{equation}
    ATE_{micro}=d_{h_{icl}/h_a'}-d_{h_q/h_a}
\end{equation}
We attribute accuracy $acc$ or ICL-time question-answer distance $d_{h_{icl}/h_a'}$ to the linearly weighted binary variables $(b,e)$ or zero-shot distance $d_{h_q/h_a}$, weight analysis is relevant to $ATE$.

\end{document}